%% file: iclr2025_conference.tex
\documentclass{article} 
\usepackage{iclr2025_conference,times}

\input{math_commands.tex}

\usepackage[pdftex]{graphicx}
\usepackage{hyperref}
\usepackage{url}
\usepackage{enumitem}
\usepackage{multirow}
\usepackage{wrapfig}
\def\papername{X-NeMo}

\usepackage{cuted}
\usepackage{capt-of}

\title{\papername: Expressive Neural Motion Reenactment via Disentangled Latent Attention}

\author{Xiaochen Zhao$^{1,2}$\footnotemark[1] , Hongyi Xu$^{2}$ , Guoxian Song$^{2}$, You Xie$^{2}$, Chenxu Zhang$^{2}$, Xiu Li$^{2}$, \\ 
\textbf{Linjie Luo$^{2}$, Jinli Suo$^{1}$, Yebin Liu$^{1}$\footnotemark[2]} \\
$^{1}$ Tsinghua University, $^{2}$ ByteDance Inc. \\ 
}

{
  \renewcommand{\thefootnote}%
    {\fnsymbol{footnote}}
  \footnotetext[1]{Work done during the internship at ByteDance.}
  \footnotetext[2]{Corresponding author.}
}

\iclrfinalcopy 
\begin{document}

\maketitle

\begin{figure}[h]
  \begin{center}
  \includegraphics[width=1.0\textwidth]{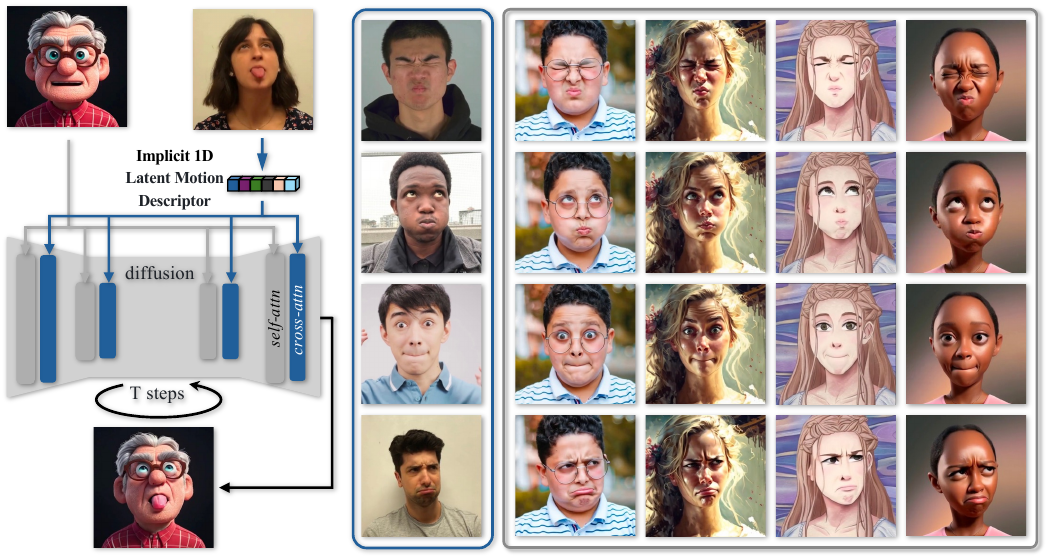}
  \end{center}
  \caption{We present~\papername, a diffusion-based portrait animation framework that integrates expressive 1D latent motion descriptors with identity-disentangled motion control through cross-attention mechanisms (left). Our method enables meticulous transfer of expressive head poses and detailed facial expressions while maintaining identity consistency, even across subjects with distinct appearances, styles and facial structures (right).
}
  \label{fig:teaser}
\end{figure}


\begin{abstract}
We propose~\papername, a novel zero-shot diffusion-based portrait animation pipeline that animates a static portrait using facial movements from a driving video of a different individual. 
Our work first identifies the root causes of the key issues in prior approaches, such as identity leakage and difficulty in capturing subtle and extreme expressions. To address these challenges, we introduce a fully end-to-end training framework that distills a 1D identity-agnostic latent motion descriptor from driving image, effectively controlling motion through cross-attention during image generation. Our implicit motion descriptor captures expressive facial motion in fine detail, learned end-to-end from a diverse video dataset without reliance on pretrained motion detectors.  We further enhance expressiveness and disentangle motion latents from identity cues by supervising their learning with a dual GAN decoder, alongside spatial and color augmentations. By embedding the driving motion into a 1D latent vector and controlling motion via cross-attention rather than additive spatial guidance, our design eliminates the transmission of spatial-aligned structural clues from the driving condition to the diffusion backbone, substantially mitigating identity leakage. Extensive experiments demonstrate that~\papername~surpasses state-of-the-art baselines, producing highly expressive animations with superior identity resemblance. Our \href{https://github.com/bytedance/x-nemo-inference}{code and models} are available for research. 
\end{abstract}

\input{sections/intro}
\input{sections/related_work}
\input{sections/method}
\input{sections/experiment}

\bibliography{iclr2025_conference}
\bibliographystyle{iclr2025_conference}

\appendix
\input{sections/supp}

\end{document}

%% file: math_commands.tex
\usepackage{amsmath,amsfonts,bm}

\def\eqref#1{equation~\ref{#1}}

\def\1{\bm{1}}

\DeclareMathAlphabet{\mathsfit}{\encodingdefault}{\sfdefault}{m}{sl}
\SetMathAlphabet{\mathsfit}{bold}{\encodingdefault}{\sfdefault}{bx}{n}



%% file: sections/intro.tex
\section{Introduction}
We investigate the task of portrait animation, where a static portrait is animated using head movements and facial expressions derived from a driving video of a different subject. This task has garnered growing interest owing to its versatile applications in video conferencing, visual effects and digital agents. Building on prior research,  we aim to advent the field of zero-shot portrait reenactment by synthesizing \emph{highly expressive} animations while maintaining \emph{identity resemblance} to the reference portraits  with minimal loss.

Commencing with the pioneering works~\cite{siarohin2019first,siarohin2019animating}, portrait animation has primarily involved extracting motion features from a driving video followed with a generative process, such as GANs~\cite{goodfellow2014generative,karras2019style,Karras2020stylegan2} or diffusion models~\cite{ho2020denoising,song2020denoising,song2020score,rombach2022high}, conditioned on the reference appearance and derived motion features. Recent advancements in diffusion models have achieved unprecedented diversity and quality in image generation, prompting us to utilize their generative capabilities \cite{saharia2022photorealistic,sd2022} for portrait animation. Recent approaches have tackled portrait animation as a controlled image-to-video diffusion task, where the reference appearance is cross-queried through mutual self-attention~\cite{cao2023masactrl} whereas the driving motion signal is integrated into the denoising process using frameworks like ControlNet~\cite{zhang2023adding} or lighter-weight PoseGuider~\cite{hu2023animateanyone}. The driving motion is represented either through explicit semantic signals such as facial landmarks~\cite{ma2024follow,wei2024aniportrait,chang2024magicpose}, dense pose~\cite{xu2024magicanimate} and facial template renderings~\cite{chen2024anifacediff}, or through implicit motion features learnt from synthetic cross-identity image pairs with aligned expression but different identities~\cite{xie2024x,yang2024megactor}.  
Despite significant progress in realism and dynamics, these diffusion-based methods still struggle to capture extreme or subtle expressions and often suffer from identity drifting, particularly when the reference and driving identities differ substantially.

We identify two main factors contributing to the challenges in expressiveness and identity resemblance in prior network designs. First, explicit motion descriptors like facial landmarks or blendshapes, are often too coarse to capture extreme or subtle facial motions and rely heavily on the robustness and accuracy of external motion detectors.  Although these descriptors do not contain RGB appearance, they encode the facial structure of the driving identity, leading to undesirable identity leakage in cross-identity animations. Recent approaches ~\cite{xie2024x,yang2024megactor} have attempted to derive implicit motion signals directly from synthetic cross-identity training image pairs generated using an off-the-shelf portrait animator (e.g.,~\cite{wang2021one}).  Despite substantial improvements in expressiveness and stability, these methods remain constrained by the capacity of pretrained portrait animators which struggle with complex expressions (e.g., tongue protrusion, cheek puffing). Additionally, sharing aligned facial structures in training pairs  inadvertently pass identity information onto the learnt implicit motion features. Second, prior diffusion-based approaches often guide motion control using spatially-aligned 2D conditions via ControlNet or PoseGuider. 
While effective for self-driven motion, this approach encourages the diffusion backbone to take a shortcut to mimic the 2D layout rather than fully leveraging semantic mappings between reference and driving images, leading to identity leakage during expression transfer across different subjects.

In this work, we propose \papername, a novel portrait animation framework that enables \emph{end-to-end} self-supervised learning of a compact 1D latent motion descriptor, facilitating effective motion control in diffusion models via \emph{cross attentions}. Specifically, we introduce a motion encoder to extract a 1-D \emph{identity-agnostic} motion latent from the original driving image, and modulate this controlling motion descriptor into the diffusion backbone via cross-attentions. By training end-to-end with the image generator, our encoder fully leverages the motion diversity and richness embedded in our training video collections, without reliance on off-the-shelf motion detectors. 
We restrict the dimensionality of the latent embedding, functioning as a low-pass filter~\cite{burkov2020neural}, and format it as a 1D global motion descriptor that excludes 2D structural cues from the driving image. Furthermore, by using cross motion attentions rather than spatial additive guidance, we ensure that the backbone remains agnostic to the identity structural signals from the motion control branch. This structure-agnostic motion control enables various augmentations like color jittering and spatial transformations, promoting the self-supervised disentanglement of identity and motion. In addition to the diffusion loss, we incorporate a dual GAN-based decoder head and refine the learning of our motion latent space with image-level losses that capture subtle and detailed expressions. 
Our design effectively mitigates the aforementioned shortcut learning, and compels the network to interpret fine-grained motion semantics during both motion encoding and image generation stages.

Trained on a collective of public video datasets~\cite{zhang2021flow,xie2022vfhq,kirschstein2023nersemble}, our method excels at faithfully capturing both extreme and nuanced facial motions and transferring them across subjects even with distinct identity attributes. We extensively evaluate our model across our challenging benchmarks and~\papername~outperforms state-of-the-art portrait animation baselines both quantitatively and qualitatively. Additionally, our expressive latent motion descriptor serves as a unified identity-agnostic embedding, facilitating motion interpolation and video outpainting applications beyond portrait animation. 
We summarize our contributions as follows,
\begin{itemize}[nolistsep,leftmargin=*]
\setlength{\itemsep}{3pt}
\setlength{\parskip}{0pt}
\setlength{\parsep}{0pt}
\item A novel diffusion-based portrait animation pipeline, trained fully end-to-end, achieving state-of-the-art performance in terms of motion accuracy, expressiveness and identity disentanglement. 
\item A structure-agnostic motion control scheme that learns a 1-D identity-disentangled latent motion descriptor and modulates control into image generation via cross-attentions, effectively addressing the long-standing issues of identity entanglement and motion expressiveness loss.  
\item A set of carefully designed strategies during both training (e.g., dual head latent supervision, augmentations and reference feature masking) and inference that substantially enhance the model performance, as supported by extensive ablation studies. 
\item Demonstration of captivating zero-shot portrait animations and generations.
\end{itemize}

%% file: sections/related_work.tex
\section{Related Works}

\paragraph{GAN-based Portrait Animation.} Video-driven face reenactment seeks to accurately transfer facial expressions and head movements from a driving video to a target image. Common approaches have primarily leveraged Generative Adversarial Networks (GANs)~\cite{goodfellow2014generative,karras2019style,Karras2020stylegan2} to model and capture the intricate motion dynamics between source and target identities. Broadly, these methods can be categorized into two classes: The first class is based on \textit{explicit motion representations}~\cite{siarohin2019animating,siarohin2019first,ren2021pirenderer,wang2021one,mallya2022implicit,yin2022styleheat,gao2023high,doukas2021headgan,guo2024liveportrait,zhao2022thin}, such as 3D face model parameters, landmarks, or latent keypoints, which use structured information to disentangle appearance and motion, but struggle with large pose changes or dynamic expressions. 
The second category involves \textit{latent motion representations}~\cite{burkov2020neural,liang2022expressive,zhou2021pose,pang2023dpe,wang2022latent,wang2023progressive,drobyshev2022megaportraits,drobyshev2024emoportraits,xu2024vasa}, embedding motion information in a latent space, offering improved expressiveness but relying on complex loss functions and hyperparameters to achieve identity-motion disentanglement. 
While more effective at transferring subtle expressions, such methods are still limited by the capability of GAN-based generators in handling extreme expressions and out-of-domain portrait styles. 
Our work follows this disentangled representation learning approach, but instead we use a Diffusion Model as the generator, which offers significantly improved generation capabilities with diverse and complex portrait styles. 

\paragraph{Diffusion-based Portrait Animation.} Diffusion models~\cite{ho2020denoising,song2020denoising,song2020score}  have demonstrated strong generative capabilities, with Latent Diffusion Models (LDMs)~\cite{rombach2022high} further advancing its efficiency by operating in a lower-dimensional latent space.  
Recent works~\cite{liu2024towards,xu2024facechain,han2023generalist,varanka2024towards,paskaleva2024unified} have explored adapting pre-trained LDMs~\cite{sd2022} for conditional portrait generation, by mapping reference images and driving signals into the text embeddings (e.g., using CLIP~\cite{khandelwal2022simple}) and injecting them into cross-attention layers.  
While effective for coarse-level facial expression editing, these methods still struggle with appearance and motion consistency in portrait video animation.~\cite{hu2023animateanyone,xu2024magicanimate,chang2024magicpose} designed for human body animation have shown that the combination of a dual U-Net with mutual self-attention~\cite{cao2023masactrl} and temporal module~\cite{guoanimatediff} is able to maintain motion smoothness with consistent appearance. This framework has been extended to portrait animation in several works~\cite{tian2024emo,xie2024x,wei2024aniportrait,xu2024hallo,yang2024megactor,wang2024v,ma2024follow,chen2024echomimic}, often using ControlNet~\cite{zhang2023adding} or PoseGuider~\cite{hu2023animateanyone} for motion control. 
During training, they rely on explicit representations like 
facial keypoints~\cite{ma2024follow,wei2024aniportrait}, facial mesh renderings~\cite{chen2024anifacediff}, or synthetic cross-identity portraits~\cite{xie2024x,yang2024megactor}.
In contrast, our method learns a latent motion representation end-to-end with our diffusion backbone, and incorporates motion control with cross-attentions, effectively preventing identity leakage.   

%% file: sections/method.tex
\begin{figure}[t]
  \begin{center}
  \includegraphics[width=1.0\textwidth]{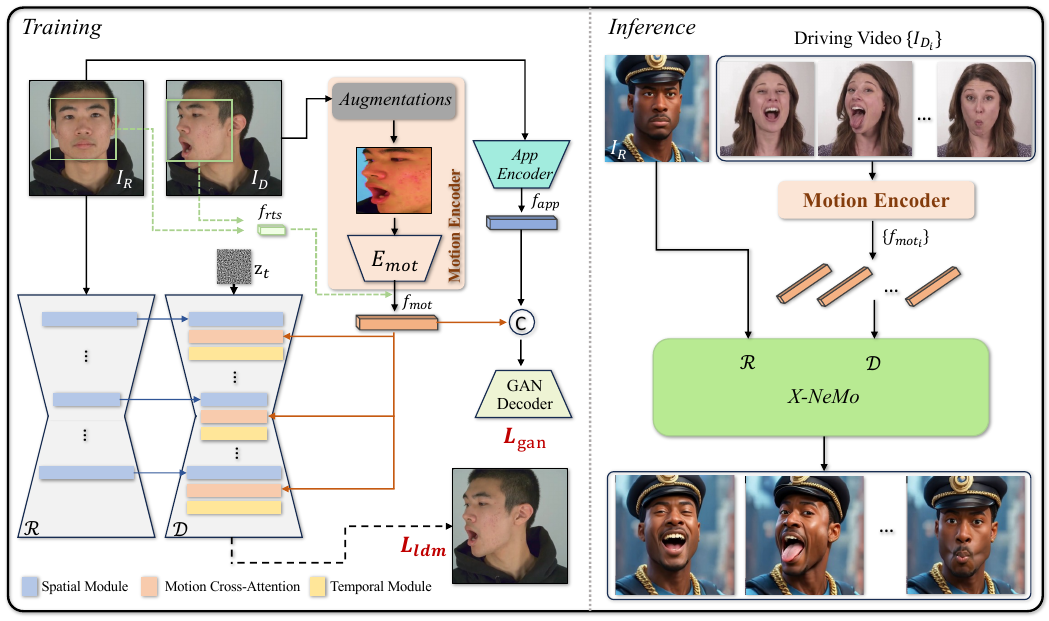}
  \end{center}
  \vspace{-2mm}
  \caption{Overview of~\papername.  We leverage a pretrained diffusion model $\mathcal{D}$ as the rendering backbone and incorporate a reference network module $\mathcal{R}$ for appearance conditioning, along with temporal modules for cross-frame consistency. For motion control, we train a latent motion embedding $f_{mot}$ encoded from the driving image $I_D$ after applying spatial and color augmentations. Alongside the relative translation and scaling $f_{rts}$ of the face bounding box from reference $I_R$ and driving image $I_D$, we integrate the latent motion conditions into the diffusion backbone using newly inserted cross-attention layers. Besides the original diffusion loss $L_{ldm}$, we supervise the learning of our latent motion embedding with a jointly trained GAN decoder head using image-level losses $L_{gan}.$ During inference, we derive the latent motion codes directly from each driving frame, allowing us to synthesize expressive and precise animations while strictly maintain identity resemblance to the reference image. 
}
\vspace{-4mm}
  \label{fig:pipe}
\end{figure}

\vspace{-3mm}
\section{Method}
Given a single portrait as the reference image $I_{R}$, our objective is to generate a head animation sequence $\{I_{R->D_i}\}$ of length $l$, conditioned on a driving video $I_{D_i}$, where $i=1,\ldots,l$ denotes the frame index. The generated frames  $\{I_{R->D_i}\}$ aim to preserve the identity features and background content depicted in $I_{R}$ while accurately replicating the head pose and facial expressions featured in each corresponding driving frame $I_{D_i}$. While portrait animation algorithms are generally trained as a frame reconstruction task over video datasets, the $I_R$ and $I_D$ may feature distinct identities during inference, enabling cross-identity motion transfer.

For our task, we harness the generative capabilities of pre-trained Latent Diffusion Models~\cite{sd2022} for image generation. Although our method shares some network modules with prior diffusion-based approaches (Section~\ref{sec:prelim}), it innovates on motion control by addressing the root causes behind the loss of expressiveness and identity resemblance. We  introduce our fully end-to-end learning framework that achieves fine-grained, identity-agnostic motion control through cross-attention to a co-learned implicit motion descriptor (Section~\ref{sec:framework}). To assist the self-supervised learning of motion and identity disentanglement, we present a set of carefully designed training strategies (Section~\ref{sec:strategies}). Figure~\ref{fig:pipe} provides an overview of our training and inference pipeline. 

\subsection{Preliminaries}
\label{sec:prelim}

\vspace{-2mm}
\paragraph{Latent Diffusion Model.} Facilitated by a pretrained auto-encoder, latent diffusion models~\cite{rombach2022high} are a class of diffusion models~\cite{ho2020denoising,song2020denoising,song2020score} that synthesize desired samples in the image latent space, starting from Gaussian noise $z_T \sim N(0,1)$ and refining through $T$ denoising steps. During training, latent representations of images are progressively corrupted by Gaussian noise $\epsilon$, following the Denoising Diffusion Probabilistic Model (DDPM) framework~\cite{ho2020denoising}. A UNet-based denoising backbone network $\mathcal{D}$ containing intervened layers of convolutions and self-/cross-attentions, is trained to learn the reverse denoising process. 

\vspace{-2mm}
\paragraph{Portrait Animation.} Recently a line of work~\cite{tian2024emo,xie2024x,wei2024aniportrait,xu2024hallo,yang2024megactor,wang2024v,ma2024follow,chen2024echomimic} have explored leveraging the generative power of pretrained LDM, such as Stable Diffusion~\cite{sd2022}, for portrait animation. While exhibiting slight algorithmic variations, these methods generally employ similar components to transfer driving motions onto the reference image. Specifically, a reference network $\mathcal{R}$ ~\cite{cao2023masactrl}, sharing the same architecture with the UNet $\mathcal{D}$, extracts reference features of identity appearance and background which are then cross-queried by the UNet self-attention blocks. 
Motion control is achieved through an additional module,  often formatted in ControlNet~\cite{zhang2023adding} or lighter-weight PoseGuider~\cite{hu2023animateanyone},  translating driving conditions into 2D spatially-aligned offsets additive to the UNet features. To maintain consistency across animated frames, temporal modules~\cite{guo2023animatediff}, which incorporates cross-frame attentions,  are intervened with the spatial transformer blocks. 

While effective to some extent, prior methods often fall short in expressiveness and suffer from identity leakage in the generated animations. First, expressiveness is limited by the coarse granularity of the driving motion conditions, such as facial landmarks or synthetic training images~\cite{xie2024x}, which fail to capture complex and subtle expressions like frowning or puckering. 
Second, 
while prior approaches mostly address appearance leakage, they overlook the leakage of 2D facial structure and spatial layout embedded in the driving conditions, whether through landmarks or synthesized images.
ControlNet-like mechanisms transform these motion conditions into spatially-aligned offsets within the UNet’s intermediate features. 
This reliance on spatial alignment causes the UNet to bypass the need to interpret semantic correspondences between the reference and driving faces, resulting in undesirable identity drift during cross-identity animation at inference.

\subsection{Pipeline with Identity-Disentangled Implicit Motion Control}
\label{sec:framework}

As shown in Figure~\ref{fig:pipe}, we follow the existing UNet-based latent diffusion framework~\cite{sd2022}, integrating both the reference network and temporal modules. However, our key innovation lies in a novel motion control module, designed to tackle challenges in motion expressiveness and identity consistency, particularly during cross-identity reenactments. A core design principle of our approach is to distill motion directly from the original driving images, while ensuring  the image generation backbone operates independently of any appearance or structural clues from the motion control path.

\vspace{-3mm}
\paragraph{Latent Motion Descriptor.}  For motion extraction, we employ an image encoder $E_{mot}$, to learn an implicit latent representation, $f_{mot}$, that captures facial motions across varying levels of granularity. Similar to the approaches in~\cite{wang2022latent,wang2023progressive}, we formulate the motion latent representation $f_{mot}$ as a \emph{low-dimensional 1-D global} descriptor. The motion encoder $E_{mot}$ consists of intervened layers of convolution-based feature extraction and self-attention, followed by MLP layers, which encode the motion into a 1D latent vector, 
thereby eliminating spatial positional information (i.e., image structure) along the encoding process.  Following the information bottleneck principle~\cite{tishby2000information}, we employ a larger network capacity (i.e., the reference net $\mathcal{R}$) and higher feature dimensions (i.e., multi-scale feature maps) for appearance modeling, while using a smaller network capacity ($E_{mot}$) and lower feature dimension ( $f_{mot}$) for motion encoding.  
This design, functioning as a low-pass bottleneck filter, encourages the emergence of disentangled representations that effectively capture key semantics of facial motion without entangling with appearance information. Furthermore, unlike previous methods that rely on pretrained motion extractor as the driving conditions (e.g., facial landmarks), our latent motion representation is continuously optimized during the end-to-end training process. As a result, this allows our model to progressively learn and refine the motion distribution as the diffusion model is trained on more diverse and expressive video data like NerSemble~\cite{kirschstein2023nersemble}. With that, our approach enhances the expressiveness of the generated animations, as the model adapts to more complex and nuanced facial motions.

\begin{figure}[t]
  \centering
  \includegraphics[width=1.0\textwidth]{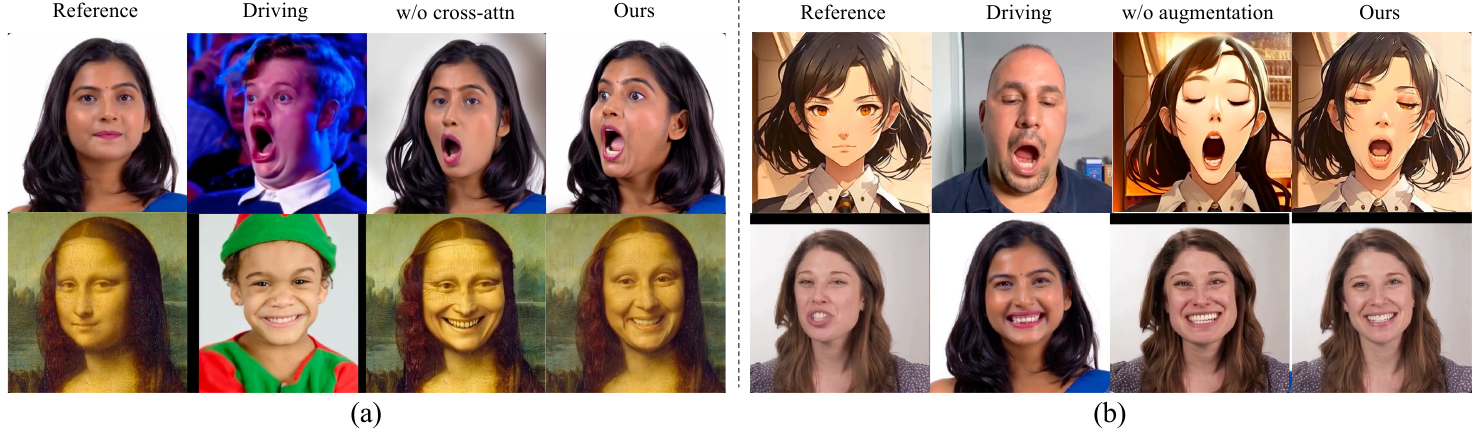}
  \vspace{-2mm}
  \caption{
  Qualitative ablation study on factors affecting identity consistency. (a) Replacing our motion cross-attentions with a control module using spatially additive guidance leads to severe leakage of the driving identity's facial structure. (b) Training without our color and spatial augmentations results in noticeable appearance leakage and identity drift.
}
\vspace{-3mm}
  \label{fig:abl_merged_idleak}
\end{figure}
\vspace{-3mm}
\paragraph{Cross-Attention Control.} To exert motion control on UNet using our latent motion descriptor, one possible approach would be to use a ControlNet-like module to guide the denoising process after transforming the 1D latent code $f_{mot}$ into a 2D spatially-aligned control map via a StyleGAN-like decoder. However, this would contradict our design goal for identity disentanglement.(Figure~\ref{fig:abl_merged_idleak}(a)) Since $f_{mot}$ is intentionally free of 2D structural information,  transforming it into a spatial control map demands additional input regarding the reference identity’s structure, thereby violating our principle that the motion control path should remain agnostic to identity-specific features. Instead the UNet should resort to the reference net for relevant identity-related information.  

Instead, we adopt a cross-attention conditioning mechanism, which has proven effective across various control modalities~\cite{rombach2022high,tian2024emo,ruiz2023dreambooth}. This allows direct injection of the latent motion embedding into the UNet without adding spatial bias.  Specifically, we insert motion-attention layers performing cross attention with the latent motion code $f_{mot}$ after each spatial transformer blocks in the backbone. This cross-attention scheme integrates the 1D motion embedding globally into the generation process, encouraging the UNet to interpret the motion condition and establish semantic correspondences between the reference and driving identity.

\subsection{Training Strategies}
\label{sec:strategies}
For training, we randomly sample two distinct frames from a video as the reference $I_{R}$ and driving $I_{D}$ image, respectively. The model is then trained to denoise the latent map of the target image $I_D$ at timestep $t,$  with the diffusion loss defined as follows,
\begin{equation}
    L_{ldm}=\mathbb{E}_{z_t,\epsilon\sim\mathcal{N}(0,1),t}\left[\|\epsilon-\epsilon_\theta(z_t,c_{ref},f_{mot})\|_2^2\right],
\end{equation}
where $\epsilon_\theta$ denotes the trainable parameters in the backbone $\mathcal{D}$, and $f_{mot}$, $c_{ref}$ represent the driving motion and reference features, respectively, extracted by $E_{mot}$ and $\mathcal{R}$. 
The training process is structured into three stages. 
The first stage is the image pretraining stage, where the backbone UNet and reference net are taken into training. The second stage involves jointly optimizing the motion encoder  $E_{mot}$ and the newly integrated motion-attention layers, forming an end-to-end encoder-generator structure. Lastly, we train the temporal modules to ensure cross-frame coherence.

However, straightforward self-supervised training of the entire framework does not inherently disentangle identity from facial motion. 
The UNet may inadvertently reconstruct the target image by borrowing appearance features from the driving image or encoding identity information into the latent motion descriptor $f_{mot}$. 
Additionally, when  $I_R$ and $I_D$ share similar expressions, the model may distill motion signals from the reference image
, hindering independent control over facial motion and identity, particularly in cross-identity reenactments. To address these issues, we propose several training strategies to fully leverage the potential of our network design in Section~\ref{sec:framework}.

\vspace{-2mm}
\paragraph{Color/Spatial Augmentation.} To suppress identity information leakage from the motion control branch, we reduce the appearance and structural consistency between the driving and target images using both color and spatial augmentations.  Specifically, 
we apply color jittering, random scaling within $30\%$, and piecewise affine transformations, to the driving image $I_D,$ altering the facial appearance and shape while preserving motion semantics. We also perform face-centered cropping to enhance spatial disparity between the driving and target, promoting the motion encoder $E_{mot}$ to focus on the facial movements and capture nuanced expressions. As shown in Figure~\ref{fig:abl_merged_idleak}(b), these augmentations effectively guide the motion encoder $E_{mot}$ towards learning identity-agnostic motion representation. To account for the disrupted head translation due to face-centered cropping, we construct a triplet $f_{rts} = (\Delta x / s_{r}, \Delta y / s_{r}, s_{d} / s_{r})$, where $(\Delta x, \Delta y)$ denotes the 2D relative distance between the face centers in the $I_{R}$ and $I_{D}$, and $ s_{d} / s_{r}$ reflects the change in bounding box scale. This triplet is processed through fully connected layers and fused with latent motion embedding $f_{mot}$. 

\begin{figure}[t]
  \centering
  \includegraphics[width=1.0\textwidth]{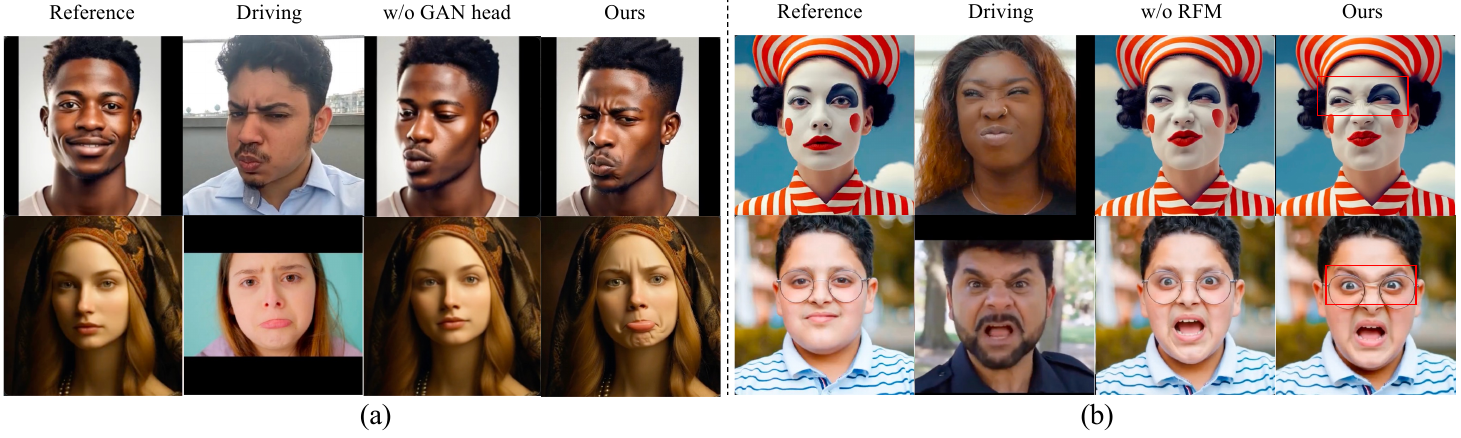}
  \caption{Qualitative ablation study on factors influencing motion expressiveness.(a) Without the dual GAN head, training solely with the diffusion loss hinders the motion encoder's ability to learn detailed and local motion patterns. (b) Our reference feature masking (RFM) strategy facilitates the transfer of fine-level facial expressions, such as the wrinkles at the nasal region.
}
\vspace{-2mm}
  \label{fig:abl_merged_expr}
\end{figure}
Owing to our design of motion control via latent motion embedding and cross-attentions, we substantially improves facial motion disentanglement from identity structure through spatial augmentations. In contrast, applying these augmentations with a ControlNet-like mechanism, which relies heavily on aligned spatial control signals, would degrade both robustness and accuracy.

\paragraph{Dual-Head Latent Supervision.} In our early experiments, we observe that our end-to-end motion control training, while effective in capturing coarse facial motions, converges slowly and struggles to depict subtle and fine-grained motions like frowning and puckering (see Figure~\ref{fig:abl_merged_expr}(a)). The latent motion embedding, acting as a low-pass filter, tends to model low-frequency movements first. Additionally, the diffusion loss used during training assigns equal weight to every ``pixel’’ in the latent noise, leading the model to prioritize a smooth motion space over capturing local, detailed expressions.  To address this, we introduce a dual GAN-based head to guide the learning of the latent motion embedding, enhancing the model's attention to fine-grained facial expressions.

Following~\cite{burkov2020neural,wang2022latent,wang2023progressive}, we employ a convolutional feature extractor network to encode the reference image into an appearance latent embedding $f_{app}$. Together with our motion latent code $f_{mot}$, these embeddings modulate a StyleGAN generator~\cite{Karras2020stylegan2} (i.e., the GAN-head decoder) to generate an RGB image, co-trained with our diffusion-based motion control. Its training losses, collectively denoted as $L_{gan}$, are formulated in image space, including a weighted $L_1$ reconstruction loss, adversarial loss, feature matching loss~\cite{burkov2020neural}, and VGG perceptual losses~\cite{simonyan2014very,Cao18vggface}. Focused on structural variations more than pixel-wise differences, these image-level losses guide the latent space learning with detailed and local motion modes. Furthermore, since the GAN head contains much fewer trainable parameters than the diffusion backbone, it converges faster, boostrapping the motion encoder and aiding the learning of motion attention layers under a well-distributed motion latent embedding.

\paragraph{Reference Feature Masking.} In line with our strategies for identity disentanglement in the motion control branch,  we also aim to mitigate motion leakage from the appearance reference network. When $I_R$ and $I_D$ exhibit similar expressions, even just partially,  the backbone network is likely to utilize the high-dimensional multi-scale appearance features as a ``shortcut’’ for motion reference, bypassing the intended reliance on our compact 1D latent motion descriptor.  While such motion leakage does not impede training data fitting during self-driven training, it hampers the learning of effective and expressive motion control (Figure~\ref{fig:abl_merged_expr}(b)).

Inspired by Masked Image Modeling~\cite{he2022masked}, we introduce reference feature masking to mitigate motion leakage in appearance features. Specifically, we apply $30\%$ uniform random masking to the appearance feature maps from the reference net $\mathcal{R}.$ The masked feature maps are flattened and used as reference keys and values for the self-attention layers within the UNet backbone. This balances the strength between appearance and motion signals, ensuring that subtle driving expressions are effectively transferred without being overshadowed by the reference expressions.

%% file: sections/experiment.tex
\section{Experiment}
\begin{figure}[t]
  \begin{center}
  \includegraphics[width=1.0\textwidth]{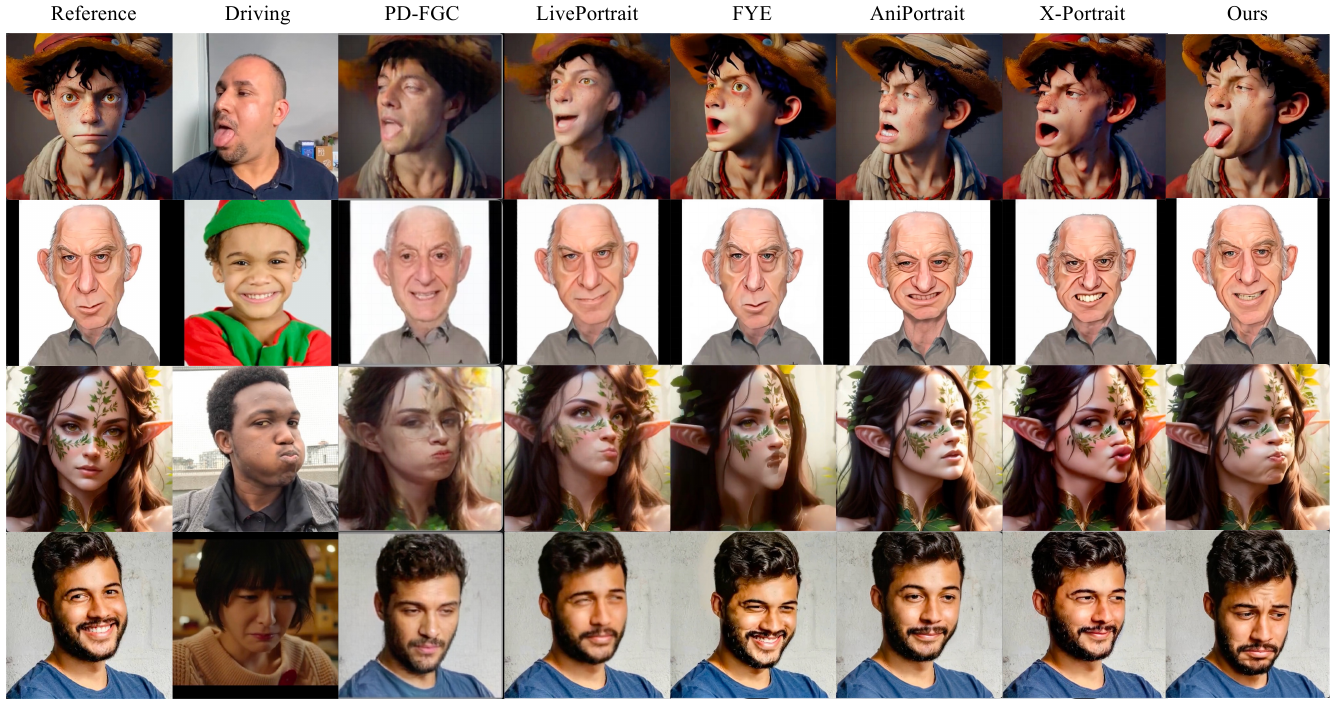}
  \end{center}
  \caption{Qualitative comparisons. Among all the methods, \papername~achieves the most accurate transfer of intricate expressions and emotional subtleties while demonstrating the highest identity resemblance, regardless of the characteristic differences between the reference and driving identities.
}
  \label{fig:abl_cmp}
  \vspace{-3mm}
\end{figure}
\subsection{Implementation Details}

We train our model on a combination of talking head datasets (HDTF~\cite{zhang2021flow}, VFHQ~\cite{xie2022vfhq}) and facial expression dataset (NerSemble~\cite{kirschstein2023nersemble}), uniformly processed at 25 fps and cropped to a $512\times512$ resolution. The training is conducted on 8 Nvidia A100 GPUs using the AdamW optimizer~\cite{yao2021adahessian} with a learning rate of $1e-5$. We use a batch size of 64 for appearance and motion control training, and a batch size of 16 for the temporal module using 24-frame video sequences. During inference, we implement the prompt traveling technique~\cite{tseng2022edge} to enhance temporal smoothness in long video generation.

For evaluation, we compile a benchmark of 100 in-the-wild reference portraits~\cite{deviantart,midjourney,pexels}, representing a broad spectrum of facial structures, appearances and styles. Additionally we collect 100 test videos from DFEW~\cite{jiang2020dfew} featuring emotionally expressive clips, alongside 200 licensed videos showcasing a diverse range of emotions, head poses, and facial expressions. Please also refer to our supplementary video for more results.

\subsection{Evaluations and Comparisons}

In our evaluation, we compare our method against  state-of-the-art video-driven portrait animation baselines, including X-Portrait~\cite{xie2024x},  AniPortrait~\cite{wei2024aniportrait},  Follow-your-Emoji (FYE)~\cite{ma2024follow}, and Echomimic~\cite{chen2024echomimic}. We also assess recent non-diffusion-based methods, including PD-FGC~\cite{wang2023progressive} that employs latent motion representation, and LivePortrait~\cite{guo2024liveportrait} which uses implicit neural landmarks. EmoPortraits~\cite{drobyshev2024emoportraits}, a GAN-based expressive portrait animation method, is excluded from our comparisons due to the lack of inference code. For fair comparisons, we finetune AniPortrait, X-Portrait, and PD-FGC using our dataset, while utilizing the released pretrained models for the remaining methods, given the unavailability of their training code. We assess performance in both self and cross reenactments, with all metrics computed at a resolution of $256\times256$ (the resolution at which PD-FGC was trained).

\vspace{-3mm}
\paragraph{Self Reenactment.}
For each test video, we utilize the first frame as the reference image, generating the entire sequence with subsequent frames acting as both the driving image and the ground truth target. We evaluate the performance by computing the L1, structural (SSIM), and perceptual (LPIPS) image losses to assess both image quality and motion accuracy. Our method, \papername, consistently outperforms all baseline methods, as shown in our numerical comparisons (Table~\ref{tab:cmp}).

\vspace{-3mm}
\paragraph{Cross Reenactment.}
Our method empowers the creation of captivating and expressive animations across diverse portraits even driven by in-the-wild videos with distinct identity features (Figure~\ref{fig:teaser},~\ref{fig:abl_merged_idleak},~\ref{fig:abl_merged_expr},~\ref{fig:abl_cmp}). Our qualitative comparisons (Figure~\ref{fig:abl_cmp}) demonstrate that~\papername~surpasses all the baselines by a significant margin in identity similarity, expression accuracy and perceptual quality. GAN-based baselines suffer from blurriness and distortions under large head motions and when applied to out-of-domain portraits.  For both exaggerated (e.g., cheek puffing, sticking out the tongue) and subtle facial expressions (e.g., biting the lip), all other methods struggle to faithfully capture and transfer these facial motion details. Additionally, our method excels in preserving identity resemblance, 
regardless of the structural difference between the reference and driving faces, while severe identity drift occurs in other diffusion-based baselines relying on spatially aligned control signals.

For quantitative assessment, given the absence of image ground truth, we employ three metrics to evaluate identity similarity, expression/head pose accuracy, and emotion consistency respectively. Specifically, we utilize the ArcFace score~\cite{deng2019arcface} to measure the cosine similarity of identity features (ID-SIM).  
Motion accuracy is calculated as the average $L_1$ difference between extracted facial blendshapes (AED) and head poses (APD) of the driving and generated images using MediaPipe~\cite{lugaresi2019mediapipe}. 
However, since blendshapes provide only a coarse motion estimation, we further employ a pretrained emotion encoder, EmoNet~\cite{toisoul2021estimation}, to assess emotion accuracy.
Specifically, we calculate the mean value of concordance correlation coefficients and Pearson correlation coefficients  for both valence and arousal to measure the emotion similarity (EMO-SIM).  
The emotion score reflect model’s performance in fine-grained expression control, as emotion recognition is highly sensitive to micro-expressions. Our method numerically surpasses all competitors, demonstrating the superior capabilities afforded by our novel motion control design (Table~\ref{tab:cmp}). 

\vspace{-3mm}
\paragraph{Applications.} 
Our latent motion descriptor acts as a unified representation for both motion comprehension and generation, supporting tasks beyond portrait animation, including (emotion-conditioned) portrait video outpainting and latent motion interpolation. With our expressive, identity-agnostic motion embedding, we are able to generate long-range expressive videos while consistently preserving the identity across diverse portraits. For more details and visual results, please refer to our supplemental paper (Section.~\ref{sec:app_app}) and accompanying video.

\begin{table}[t]
\centering
\caption{Quantitative comparison. Our method achieves superior numerical results than all the baselines in both self-driven and cross-driven reenactments. }
\begin{tabular}{l|ccc|ccc}
\hline
\multirow{2}{*}{Method} & \multicolumn{3}{c|}{Self-Reenactment} & \multicolumn{3}{c}{Cross-Reenactment}             \\ \cline{2-7} 
                        & L1$\downarrow$       & SSIM$\uparrow$      & LPIPS$\downarrow$      & ID-SIM$\uparrow$ & AED/APD$\downarrow$ & EMO-SIM$\uparrow$ \\ \hline
PD-FGC                  & 0.085  &  0.728  &   0.291   & 0.604 & 0.045/3.95 &  0.49         \\
LivePortrait            & 0.074    & 0.770     &  0.236    & 0.702 & 0.055/6.61 & 0.48         \\
X-Portrait              & 0.063    & 0.793     &  0.209     & 0.695 & 0.041/4.07 & 0.52        \\
FYE                     & 0.075    & 0.741     &  0.249   & 0.725 & 0.062/4.49 & 0.41     \\
AniPortrait             & 0.057    & 0.812     & 0.198   & 0.713 & 0.043/4.14 &  0.46       \\ \hline
Ours                    & \textbf{0.055}    & \textbf{0.826}   &  \textbf{0.168}   & \textbf{0.787} & \textbf{0.039/3.42} &  \textbf{0.65}       \\ \hline
\end{tabular}
\vspace{-3mm}
\label{tab:cmp}
\end{table}

\vspace{-2mm}
\subsection{Ablation Studies}

We ablate individual design choices by removing them from our full training pipeline. 
We validate the function of dual-head supervision in motion expressiveness by removing the GAN decoder from co-training (``w/o GAN head''). Even with extended training,  the motion encoder $E_{mot}$ struggles with detailed motions in the absence of image-level loss guidance ( Figure~\ref{fig:abl_merged_expr}(a)), as reflected by a substantial reduction in  both expression and emotion metrics (Table~\ref{tab:abl}).
We further validate the importance of end-to-end training by pretraining $E_{mot}$ with GAN losses and then freezing it while training the rest of the model sorely with the diffusion loss (``w/o e2e''). Both Table~\ref{tab:abl} and the supplemental video demonstrate that end-to-end training elicits stronger motion representation from the encoder, leveraging the diffusion model's superior generative capacity over the standalone GAN decoder.
Additionally, we assess the role of reference feature masking (``w/o RFM'') in enhancing motion accuracy. Without it, the network shows a stronger bias to the reference expressions at certain local regions(Figure~\ref{fig:abl_merged_expr}(b)), yeilding a lower emotion score (Table~\ref{tab:abl}).

\begin{wraptable}{rt}{9 cm}
\vspace{-6mm}
\caption{Quantitative ablation.}
    \centering
    \begin{tabular}{l|ccc}
    \hline
    Method                           & ID-SIM$\uparrow$ & AED/APD$\downarrow$ & EMO-SIM$\uparrow$ \\ \hline
    w/o GAN head                     & 0.789 & 0.045/4.64 & 0.43  \\
    w/o e2e          & 0.782 & 0.040/3.49 & 0.52 \\
    w/o RFM                          & \textbf{0.791} & \textbf{0.039/3.41} & 0.62 \\
    w/o augmentation                & 0.724 & 0.042/3.63 & 0.50  \\
    w/o cross-attn                & 0.697 & 0.040/3.55 & 0.48  \\ \hline
    Ours                             & 0.787 & \textbf{0.039}/3.42 & \textbf{0.65} \\ \hline
    
    \end{tabular}
    \vspace{-3mm}
\label{tab:abl}
\end{wraptable} 

We assess the efficacy of augmentation and cross-attention control in motion-identity disentanglement. When training without augmentations (``w/o augmentation''), the generations often exhibit noticeable identity leakage from the driving subject in both appearance and face structure (Figure~\ref{fig:abl_merged_idleak}(b)), as confirmed by the drop in identity similarity score (Table~\ref{tab:abl}). We also compare our method to a baseline where the motion latent is transformed into a 2D control map via an upsampling decoder and applied to the UNet using a ControlNet (``w/o cross-attn''). While effective for coarse motion control, its reliance on spatially-aligned additive controls lead to reduced identity resemblance (Figure~\ref{fig:abl_merged_expr}(a)), underscoring the necessity of our structure-agnostic motion control design.

\begin{wrapfigure}{r}{0.55\textwidth}
  \vspace{-6mm}
  \begin{center}
    \includegraphics[width=0.5\textwidth]{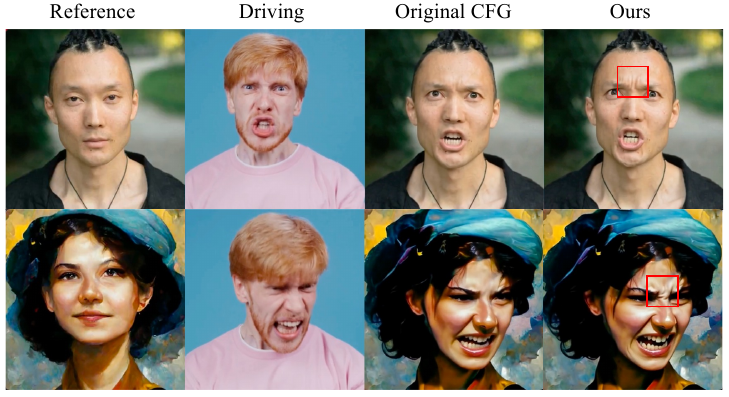}
  \end{center}
  \vspace{-5mm}
  \caption{Ablation on different CFG configurations.}
  \vspace{-2mm}
  \label{fig:abl_cfg}
\end{wrapfigure}
We leverage classifier-free guidance (CFG)~\cite{ho2022classifier} to steer the inference towards more expressive motion transfer. While straightforward for the conditional generation, we find the optimal practice by using fully masked appearance features and the motion latent $f_{ref\_mot}$ extracted from reference image $I_R$ as the negative prompts. As illustrated in Figure~\ref{fig:abl_cfg}, this CFG configuration enables the network to better distinguish between conditional appearance and motion features, facilitating more accurate and semantic motion transfer.
Our CFG is formulated as 
\begin{equation}
\tilde{\boldsymbol{\epsilon}}_\theta({z}_t,{c_{ref}},f_{mot})=(1+w)\boldsymbol{\epsilon}_\theta({z}_t,{c_{ref}},f_{mot})-w\boldsymbol{\epsilon}_\theta({z}_t, \emptyset, f_{ref\_mot}),
\end{equation}
where $w=3.5$ is the CFG scale and  $\tilde{\boldsymbol{\epsilon}}_\theta$ is the final composed noise estimate.

\section{Discussion and Conclusion}
\vspace{-2mm}
We present~\papername, a novel diffusion-based portrait animation framework that effectively disentangles motion and identity, achieving substantial improvements in generating expressive, identity-preserved animations from diverse portraits. At its core, we introduce an end-to-end learning framework that integrates latent motion representations with structure-agnostic motion control through cross-attentions, enhanced by carefully-designed training and inference strategies. We demonstrate high-quality animation results on a wide range of portraits and expressive driving videos, validating the efficacy of our approach. We believe our method offers valuable insights into the field and opens avenues for numerous downstream tasks. Code and model will be available for research. 
\vspace{-3mm}
\paragraph{Limitations and Future Work.}
\begin{wrapfigure}{r}{0.55\textwidth}
  \vspace{-6mm}
  \begin{center}
    \includegraphics[width=0.5\textwidth]{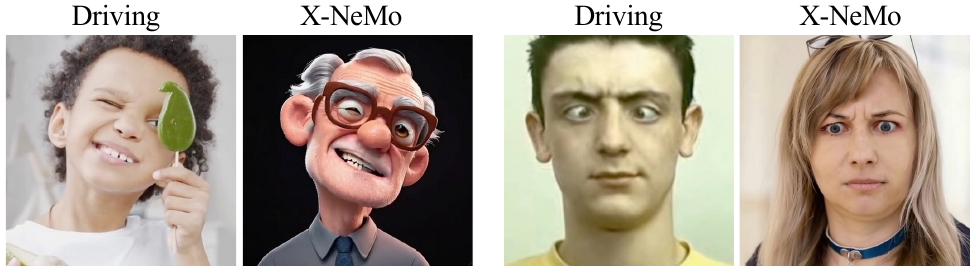}
  \end{center}
  \vspace{-5mm}
  \caption{Failure cases.}
  \vspace{-6mm}
  \label{fig:failure}
\end{wrapfigure}
Our method is trained solely on real human talking and expression videos. Consequently, out-of-domain portraits with non-human appearances, such as 3D cartoon characters, may exhibit artifacts like blurred eyes. Additionally, it  might struggle with exaggerated expressions absent from the training data (see Figure~\ref{fig:failure}). However,  our scalable end-to-end framework, free from reliance on pre-trained motion detectors, enables generalization to various styles and motions with more training data. Our motion control represents a general scheme, with which we aim to integrate into video diffusion backbones~\cite{yang2024cogvideox,opensora} in the future, for smoother and more dynamic results. 

\vspace{-3mm}
\paragraph{Ethics Statement.} 
Our work aims to improve portrait animation from a technical perspective and is not intended for malicious use. However, we recognize the potential for misuse like generating fake videos. Therefore, synthesized images and videos should clearly indicate their artificial nature.

%% file: sections/supp.tex
\section{Training and Inference Details}
\vspace{-2mm}
\paragraph{GAN Head Training Losses.}
Following ~\cite{burkov2020neural}, we train our dual GAN decoder in a self-supervised manner to reconstruct $I_{D}$ using a combination of losses. Specifically,
a $L_{1}$ reconstruction loss is employed to minimize pixel-wise $L1$ distance:
\begin{equation}
    \mathcal{L}_{recon}=\left\|I_D-I_{R\to D}\right\|_1
\end{equation}
Additionally, two perceptual losses, $L_{VGG}$ and $L_{VGGFace}$, are applied based on $L_1$ matching of ConvNet activations from a VGG-19 model~\cite{simonyan2014very} pretrained for ImageNet classification and a VGGFace model~\cite{Cao18vggface} trained for face recognition:
\begin{gather}
\mathcal{L}_{vgg}=\sum_{i=1}^{N}\|\mathrm{VGG}^i(I_{D})-\mathrm{VGG}^i(I_{R\to D})\|_1, \\
\mathcal{L}_{vggf}=\sum_{I=1}^{N}\|\mathrm{VGG_{face}}^i(I_{D})-\mathrm{VGG_{face}}^i(I_{R\to D})\|_1,
\end{gather}
where $N$ denotes the number of feature layers in each respective pre-trained VGG model.
An adversarial generative loss $L_{adv}$ is applied using a co-trained discriminator $\mathrm{D},$ while a feature matching loss $L_{fm}$  is calculated as the $L_1$ distance between discriminator
 feature maps at different layers:
\begin{equation}
\mathcal{L}_{fm}=\sum_{i=1}^{N}\|\mathrm{D}^i(I_{D})-\mathrm{D}^i(I_{R\to D})\|_1
\end{equation}
The overall learning objective for the GAN head is then formulated as:
\begin{equation}
\mathcal{L}_{gan}=\mathcal{L}_\mathrm{adv}+\lambda_r\mathcal{L}_{recon}+\lambda_{vgg}\mathcal{L}_\mathrm{vgg}+\lambda_{vggf}\mathcal{L}_\mathrm{vggf}+\lambda_{fm}\mathcal{L}_{fm}
\end{equation}
where $\lambda_r$=1.0, $\lambda_{vgg}$=3e-2, $\lambda_{vggf}$=6e-3 and $\lambda_{fm}$=10.0.

\paragraph{Inference Performance.} 
During inference, we use 25 DDIM steps~\cite{song2020denoising} with a classifier-free guidance (CFG) scale of 3.5. For generating a 1-second video at 25 frames per second, the process takes approximately 20 seconds and requires 24 GB of memory.

\paragraph{Encoder Architecture.}
$E_{mot}$ takes the classical feature alignment network~\cite{bulat2017far} as the backbone, with an additional attention layer added at both its input and output to enhance feature extraction capabilities. Finally, it outputs a 1D vector through two MLP layers. The appearance encoder for extracting $f_{app}$ is implemented as a ResNet50.

\section{More Ablations}
We provide additional visual ablations on some network and training hyperparameters.

\begin{figure}[h]
  \centering
  \includegraphics[width=0.92\textwidth]{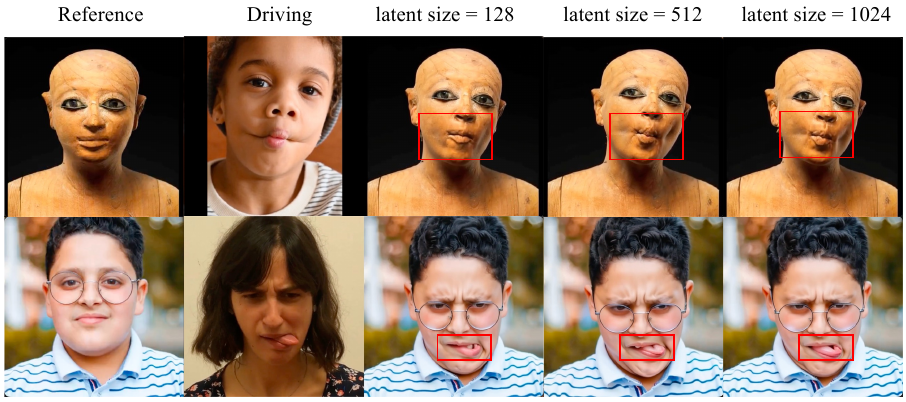}
  \caption{We ablate the effect of different sizes of latent motion embedding in capturing fine-grained intricate expressions. 
}
  \label{fig:abl_embdim}
\end{figure}

\paragraph{Motion Latent Embedding Size.} 
We evaluate the dimensionality of our motion latent embedding by comparing 128, 512, and 1024-dimensional latent codes for $f_{mot}$. In typical driving scenarios, all three configurations perform similarly in replicating facial expressions with minimal differences. However, as shown in Figure~\ref{fig:abl_embdim}, reducing the embedding size to 128 diminishes the ability to capture subtle, intricate expressions, while increasing it to 1024 provides negligible improvements. Thus, we select the 512-dimensional embedding as it balances compactness with motion expressiveness.

\begin{figure}[h]
  \centering
  \includegraphics[width=0.92\textwidth]{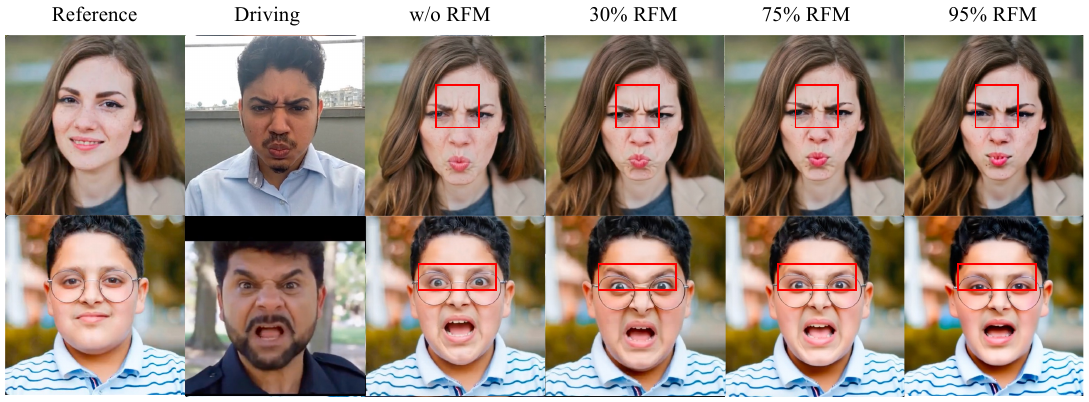}
  \caption{Qualitative comparison of different ratios of reference feature masking indicates that $30\%$ achieves the most accurate capture of driving facial motions.
}
  \label{fig:abl_rmfratio}
\end{figure}

\paragraph{Reference Feature Masking Ratio.} We assess the effectiveness of our reference feature masking strategy across different masking ratios, ranging from $0\%$, $30\%$, $75\%$, to $95\%$. As shown in Figure~\ref{fig:abl_rmfratio}, this strategy enhances the transfer of detailed facial expressions; however, excessively high masking ratios impede the model's ability to capture fine motion details and maintain identity consistency. This is likely because when the reference image's appearance is too heavily obscured during training, the motion encoder compensates by encoding appearance information, reducing the capacity of the motion embedding for expressing dynamic movements. In practice, we found that masking ratios between $20\%$ and $50\%$ achieve optimal results, with $30\%$ used in our implementation.

\section{More Results}
Please refer to our supplemental video for more expressive demo cases.

\section{Applications}
\label{sec:app_app}
Our latent motion embedding, trained end-to-end with the diffusion backbone, offers a compact, identity-agnostic, yet expressive representation for a diverse range of facial motions. Beyond the primary portrait animation task, we showcase its broader applications as a unified motion representation, enabling seamless motion interpolation, video outpainting and conditioned generation.
\begin{figure}[h]
  \centering
  \includegraphics[width=0.92\textwidth]{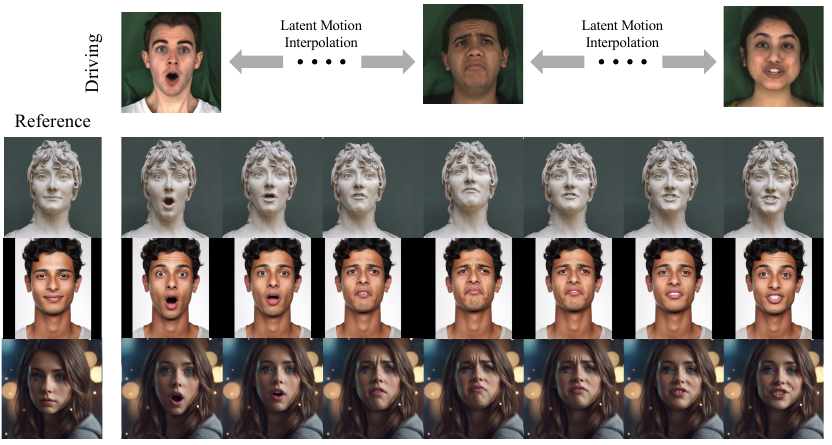}
  \caption{Latent motion interpolation. We derive latent motion codes from a few driving keyframes (top) and apply onto diverse portraits with linearly interpolated motion embeddings (bottom). 
}
  \label{fig:app_motioninter}
\end{figure}

\paragraph{Latent Motion Interpolation.} Owing to our smooth and identity-agnostic latent motion space, we are able to extract keyframe expressions from different videos, linearly interpolate the latent embeddings and apply them across diverse portraits, as showcased in Figure.~\ref{fig:app_motioninter}. This interpolation yields smooth and natural expression transitions, maintaining motion coherence across different portraits and appearance consistency with the reference images. These results underscore the robustness and identity disentanglement of our motion latent embedding.

\begin{figure}[h]
  \centering
  \includegraphics[width=0.92\textwidth]{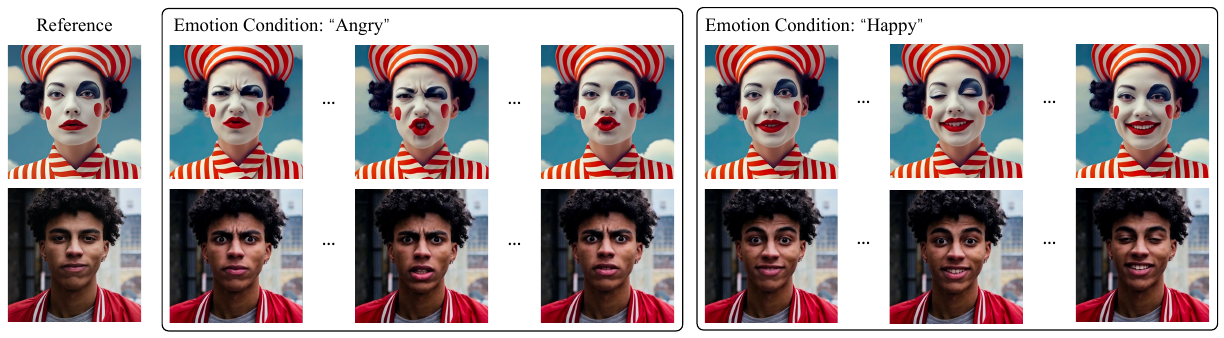}
  \caption{Emotion-conditioned portrait video generation.  
}
  \label{fig:app_emogen}
\end{figure}

\begin{figure}[h]
  \centering
  \includegraphics[width=0.92\textwidth]{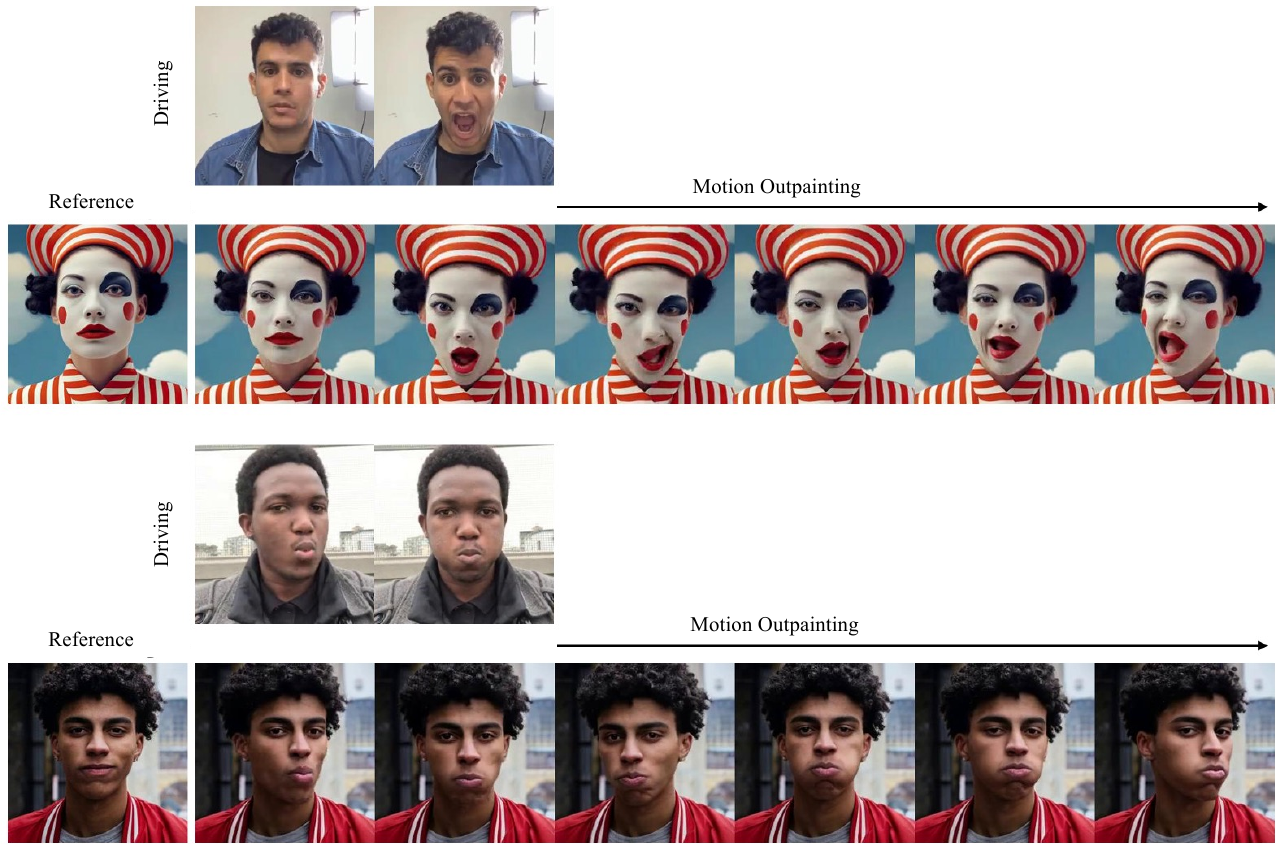}
  \caption{Portrait video outpainting. Starting from a sequence of driving motion, our model is capable of extrapolating into a long video sequence with consistent identity attributes. 
}
  \label{fig:app_motiongen}
\end{figure}

\paragraph{Portrait Video Outpainting and Generation.} By leveraging our motion latent embedding as a unified representation for motion comprehension and generation, we showcase its application in video outpainting. Specifically, we adopt the approach from T2M-GPT~\cite{zhang2023generating} to tokenize temporal latent motions by training a Vector-Quantized VAE model~\cite{esser2021taming} with a learnable codebook (4096 entries of 8-dimension code) that downsamples the temporal dimension by a factor of 4. This allows us to represent $T$ frames of motion with $T/4$ discrete motion tokens, where $T$ is the training sequence length (we use $T=128$), facilitating the use of GPT-like frameworks for long-sequence motion generation.  In Figure.~\ref{fig:app_motiongen}, we train a GPT2-small network that extends preceding motions derived from a driving video with extrapolated motions. The results show natural and expressive generated sequences, thanks to the strong representation power of our latent motion embedding. Moreover, as more facial video datasets containing multimodal annotations (e.g., text and audio) become available, our method can seamlessly extend to multimodal facial video generation within a unified framework. As an example, we illustrate emotion-conditioned portrait video generation in Figure~\ref{fig:app_emogen}, trained with MEAD dataset~\cite{wang2020mead}. 

%% file: iclr2025_conference.bbl
\begin{thebibliography}{74}
\providecommand{\natexlab}[1]{#1}
\providecommand{\url}[1]{\texttt{#1}}
\expandafter\ifx\csname urlstyle\endcsname\relax
  \providecommand{\doi}[1]{doi: #1}\else
  \providecommand{\doi}{doi: \begingroup \urlstyle{rm}\Url}\fi

\bibitem[AI(2022)]{sd2022}
Stability AI.
\newblock Stable diffusion v1.5 model card.
\newblock \emph{https://huggingface.co/runwayml/stable-diffusion-v1-5}, 2022.

\bibitem[Bulat \& Tzimiropoulos(2017)Bulat and Tzimiropoulos]{bulat2017far}
Adrian Bulat and Georgios Tzimiropoulos.
\newblock How far are we from solving the 2d \& 3d face alignment problem?(and a dataset of 230,000 3d facial landmarks).
\newblock In \emph{Proceedings of the IEEE international conference on computer vision}, pp.\  1021--1030, 2017.

\bibitem[Burkov et~al.(2020)Burkov, Pasechnik, Grigorev, and Lempitsky]{burkov2020neural}
Egor Burkov, Igor Pasechnik, Artur Grigorev, and Victor Lempitsky.
\newblock Neural head reenactment with latent pose descriptors.
\newblock In \emph{CVPR}, 2020.

\bibitem[Cao et~al.(2023)Cao, Wang, Qi, Shan, Qie, and Zheng]{cao2023masactrl}
Mingdeng Cao, Xintao Wang, Zhongang Qi, Ying Shan, Xiaohu Qie, and Yinqiang Zheng.
\newblock Masactrl: Tuning-free mutual self-attention control for consistent image synthesis and editing.
\newblock \emph{arXiv preprint arXiv:2304.08465}, 2023.

\bibitem[Cao et~al.(2018)Cao, Shen, Xie, Parkhi, and Zisserman]{Cao18vggface}
Qiong Cao, Li~Shen, Weidi Xie, Omkar~M. Parkhi, and Andrew Zisserman.
\newblock {VGGFace2}: A dataset for recognising faces across pose and age.
\newblock In \emph{International Conference on Automatic Face and Gesture Recognition}, 2018.

\bibitem[Chang et~al.(2024)Chang, Shi, Gao, Xu, Fu, Song, Yan, Zhu, Yang, and Soleymani]{chang2024magicpose}
Di~Chang, Yichun Shi, Quankai Gao, Hongyi Xu, Jessica Fu, Guoxian Song, Qing Yan, Yizhe Zhu, Xiao Yang, and Mohammad Soleymani.
\newblock Magicpose: Realistic human poses and facial expressions retargeting with identity-aware diffusion.
\newblock In \emph{ICML}, 2024.

\bibitem[Chen et~al.(2024{\natexlab{a}})Chen, Seneviratne, Wang, Hu, Saha, Hasan, Rasnayaka, Malepathirana, Gong, and Halgamuge]{chen2024anifacediff}
Ken Chen, Sachith Seneviratne, Wei Wang, Dongting Hu, Sanjay Saha, Md~Tarek Hasan, Sanka Rasnayaka, Tamasha Malepathirana, Mingming Gong, and Saman Halgamuge.
\newblock Anifacediff: High-fidelity face reenactment via facial parametric conditioned diffusion models.
\newblock \emph{arXiv preprint arXiv:2406.13272}, 2024{\natexlab{a}}.

\bibitem[Chen et~al.(2024{\natexlab{b}})Chen, Cao, Chen, Li, and Ma]{chen2024echomimic}
Zhiyuan Chen, Jiajiong Cao, Zhiquan Chen, Yuming Li, and Chenguang Ma.
\newblock Echomimic: Lifelike audio-driven portrait animations through editable landmark conditions.
\newblock \emph{arXiv preprint arXiv:2407.08136}, 2024{\natexlab{b}}.

\bibitem[Deng et~al.(2019)Deng, Guo, Xue, and Zafeiriou]{deng2019arcface}
Jiankang Deng, Jia Guo, Niannan Xue, and Stefanos Zafeiriou.
\newblock Arcface: Additive angular margin loss for deep face recognition.
\newblock In \emph{CVPR}, pp.\  4690--4699, 2019.

\bibitem[DeviantArt(2024)]{deviantart}
DeviantArt.
\newblock deviantart.
\newblock \emph{https://www.deviantart.com}, 2024.

\bibitem[Doukas et~al.(2021)Doukas, Zafeiriou, and Sharmanska]{doukas2021headgan}
Michail~Christos Doukas, Stefanos Zafeiriou, and Viktoriia Sharmanska.
\newblock Headgan: One-shot neural head synthesis and editing.
\newblock In \emph{ICCV}, 2021.

\bibitem[Drobyshev et~al.(2022)Drobyshev, Chelishev, Khakhulin, Ivakhnenko, Lempitsky, and Zakharov]{drobyshev2022megaportraits}
Nikita Drobyshev, Jenya Chelishev, Taras Khakhulin, Aleksei Ivakhnenko, Victor Lempitsky, and Egor Zakharov.
\newblock Megaportraits: One-shot megapixel neural head avatars.
\newblock In \emph{ACM MM}, 2022.

\bibitem[Drobyshev et~al.(2024)Drobyshev, Casademunt, Vougioukas, Landgraf, Petridis, and Pantic]{drobyshev2024emoportraits}
Nikita Drobyshev, Antoni~Bigata Casademunt, Konstantinos Vougioukas, Zoe Landgraf, Stavros Petridis, and Maja Pantic.
\newblock Emoportraits: Emotion-enhanced multimodal one-shot head avatars.
\newblock In \emph{CVPR}, 2024.

\bibitem[Esser et~al.(2021)Esser, Rombach, and Ommer]{esser2021taming}
Patrick Esser, Robin Rombach, and Bjorn Ommer.
\newblock Taming transformers for high-resolution image synthesis.
\newblock In \emph{Proceedings of the IEEE/CVF conference on computer vision and pattern recognition}, pp.\  12873--12883, 2021.

\bibitem[Gao et~al.(2023)Gao, Zhou, Wang, Li, Ming, and Lu]{gao2023high}
Yue Gao, Yuan Zhou, Jinglu Wang, Xiao Li, Xiang Ming, and Yan Lu.
\newblock High-fidelity and freely controllable talking head video generation.
\newblock In \emph{CVPR}, 2023.

\bibitem[Goodfellow et~al.(2014)Goodfellow, Pouget-Abadie, Mirza, Xu, Warde-Farley, Ozair, Courville, and Bengio]{goodfellow2014generative}
Ian Goodfellow, Jean Pouget-Abadie, Mehdi Mirza, Bing Xu, David Warde-Farley, Sherjil Ozair, Aaron Courville, and Yoshua Bengio.
\newblock Generative adversarial nets.
\newblock In \emph{NeurIPS}, 2014.

\bibitem[Guo et~al.(2024{\natexlab{a}})Guo, Zhang, Liu, Zhong, Zhang, Wan, and Zhang]{guo2024liveportrait}
Jianzhu Guo, Dingyun Zhang, Xiaoqiang Liu, Zhizhou Zhong, Yuan Zhang, Pengfei Wan, and Di~Zhang.
\newblock Liveportrait: Efficient portrait animation with stitching and retargeting control.
\newblock \emph{arXiv preprint arXiv:2407.03168}, 2024{\natexlab{a}}.

\bibitem[Guo et~al.(2023)Guo, Yang, Rao, Wang, Qiao, Lin, and Dai]{guo2023animatediff}
Yuwei Guo, Ceyuan Yang, Anyi Rao, Yaohui Wang, Yu~Qiao, Dahua Lin, and Bo~Dai.
\newblock Animatediff: Animate your personalized text-to-image diffusion models without specific tuning.
\newblock \emph{arXiv preprint arXiv:2307.04725}, 2023.

\bibitem[Guo et~al.(2024{\natexlab{b}})Guo, Yang, Rao, Liang, Wang, Qiao, Agrawala, Lin, and Dai]{guoanimatediff}
Yuwei Guo, Ceyuan Yang, Anyi Rao, Zhengyang Liang, Yaohui Wang, Yu~Qiao, Maneesh Agrawala, Dahua Lin, and Bo~Dai.
\newblock Animatediff: Animate your personalized text-to-image diffusion models without specific tuning.
\newblock In \emph{ICLR}, 2024{\natexlab{b}}.

\bibitem[Han et~al.(2023)Han, Zhang, Zhu, Li, Ge, Li, Wang, Liu, Liu, and Tai]{han2023generalist}
Yue Han, Jiangning Zhang, Junwei Zhu, Xiangtai Li, Yanhao Ge, Wei Li, Chengjie Wang, Yong Liu, Xiaoming Liu, and Ying Tai.
\newblock A generalist facex via learning unified facial representation.
\newblock \emph{arXiv preprint arXiv:2401.00551}, 2023.

\bibitem[He et~al.(2022)He, Chen, Xie, Li, Doll{\'a}r, and Girshick]{he2022masked}
Kaiming He, Xinlei Chen, Saining Xie, Yanghao Li, Piotr Doll{\'a}r, and Ross Girshick.
\newblock Masked autoencoders are scalable vision learners.
\newblock In \emph{CVPR}, pp.\  16000--16009, 2022.

\bibitem[Ho \& Salimans(2022)Ho and Salimans]{ho2022classifier}
Jonathan Ho and Tim Salimans.
\newblock Classifier-free diffusion guidance.
\newblock \emph{arXiv preprint arXiv:2207.12598}, 2022.

\bibitem[Ho et~al.(2020)Ho, Jain, and Abbeel]{ho2020denoising}
Jonathan Ho, Ajay Jain, and Pieter Abbeel.
\newblock Denoising diffusion probabilistic models.
\newblock \emph{NeurIPS}, 33:\penalty0 6840--6851, 2020.

\bibitem[Hu et~al.(2023)Hu, Gao, Zhang, Sun, Zhang, and Bo]{hu2023animateanyone}
Li~Hu, Xin Gao, Peng Zhang, Ke~Sun, Bang Zhang, and Liefeng Bo.
\newblock Animate anyone: Consistent and controllable image-to-video synthesis for character animation.
\newblock \emph{arXiv preprint arXiv:2311.17117}, 2023.

\bibitem[Jiang et~al.(2020)Jiang, Zong, Zheng, Tang, Xia, Lu, and Liu]{jiang2020dfew}
Xingxun Jiang, Yuan Zong, Wenming Zheng, Chuangao Tang, Wanchuang Xia, Cheng Lu, and Jiateng Liu.
\newblock Dfew: A large-scale database for recognizing dynamic facial expressions in the wild.
\newblock In \emph{Proceedings of the 28th ACM international conference on multimedia}, pp.\  2881--2889, 2020.

\bibitem[Karras et~al.(2019)Karras, Laine, and Aila]{karras2019style}
Tero Karras, Samuli Laine, and Timo Aila.
\newblock A style-based generator architecture for generative adversarial networks.
\newblock In \emph{CVPR}, 2019.

\bibitem[Karras et~al.(2020)Karras, Laine, Aittala, Hellsten, Lehtinen, and Aila]{Karras2020stylegan2}
Tero Karras, Samuli Laine, Miika Aittala, Janne Hellsten, Jaakko Lehtinen, and Timo Aila.
\newblock Analyzing and improving the image quality of {StyleGAN}.
\newblock In \emph{CVPR}, 2020.

\bibitem[Khandelwal et~al.(2022)Khandelwal, Weihs, Mottaghi, and Kembhavi]{khandelwal2022simple}
Apoorv Khandelwal, Luca Weihs, Roozbeh Mottaghi, and Aniruddha Kembhavi.
\newblock Simple but effective: Clip embeddings for embodied ai.
\newblock In \emph{Proceedings of the IEEE/CVF Conference on Computer Vision and Pattern Recognition}, pp.\  14829--14838, 2022.

\bibitem[Kirschstein et~al.(2023)Kirschstein, Qian, Giebenhain, Walter, and Nie{\ss}ner]{kirschstein2023nersemble}
Tobias Kirschstein, Shenhan Qian, Simon Giebenhain, Tim Walter, and Matthias Nie{\ss}ner.
\newblock Nersemble: Multi-view radiance field reconstruction of human heads.
\newblock \emph{{ACM Transactions on Graphics}}, 2023.

\bibitem[Liang et~al.(2022)Liang, Pan, Guo, Zhou, Hong, Han, Han, Liu, Ding, and Wang]{liang2022expressive}
Borong Liang, Yan Pan, Zhizhi Guo, Hang Zhou, Zhibin Hong, Xiaoguang Han, Junyu Han, Jingtuo Liu, Errui Ding, and Jingdong Wang.
\newblock Expressive talking head generation with granular audio-visual control.
\newblock In \emph{CVPR}, 2022.

\bibitem[Liu et~al.(2024)Liu, Ma, Zhang, Hu, Fan, Lv, Ding, and Cheng]{liu2024towards}
Renshuai Liu, Bowen Ma, Wei Zhang, Zhipeng Hu, Changjie Fan, Tangjie Lv, Yu~Ding, and Xuan Cheng.
\newblock Towards a simultaneous and granular identity-expression control in personalized face generation.
\newblock In \emph{CVPR}, 2024.

\bibitem[Lugaresi et~al.(2019)Lugaresi, Tang, Nash, McClanahan, Uboweja, Hays, Zhang, Chang, Yong, Lee, et~al.]{lugaresi2019mediapipe}
Camillo Lugaresi, Jiuqiang Tang, Hadon Nash, Chris McClanahan, Esha Uboweja, Michael Hays, Fan Zhang, Chuo-Ling Chang, Ming~Guang Yong, Juhyun Lee, et~al.
\newblock Mediapipe: A framework for building perception pipelines.
\newblock \emph{arXiv preprint arXiv:1906.08172}, 2019.

\bibitem[Ma et~al.(2024)Ma, Liu, Wang, Pan, He, Yuan, Zeng, Cai, Shum, Liu, et~al.]{ma2024follow}
Yue Ma, Hongyu Liu, Hongfa Wang, Heng Pan, Yingqing He, Junkun Yuan, Ailing Zeng, Chengfei Cai, Heung-Yeung Shum, Wei Liu, et~al.
\newblock Follow-your-emoji: Fine-controllable and expressive freestyle portrait animation.
\newblock \emph{arXiv preprint arXiv:2406.01900}, 2024.

\bibitem[Mallya et~al.(2022)Mallya, Wang, and Liu]{mallya2022implicit}
Arun Mallya, Ting-Chun Wang, and Ming-Yu Liu.
\newblock Implicit warping for animation with image sets.
\newblock \emph{NeurIPS}, 2022.

\bibitem[Midjourney(2024)]{midjourney}
Midjourney.
\newblock midjourney.
\newblock \emph{https://www.midjourney.com}, 2024.

\bibitem[Pang et~al.(2023)Pang, Zhang, Quan, Fan, Cun, Shan, and Yan]{pang2023dpe}
Youxin Pang, Yong Zhang, Weize Quan, Yanbo Fan, Xiaodong Cun, Ying Shan, and Dong-ming Yan.
\newblock Dpe: Disentanglement of pose and expression for general video portrait editing.
\newblock In \emph{CVPR}, 2023.

\bibitem[Paskaleva et~al.(2024)Paskaleva, Holubakha, Ilic, Motamed, Van~Gool, and Paudel]{paskaleva2024unified}
Reni Paskaleva, Mykyta Holubakha, Andela Ilic, Saman Motamed, Luc Van~Gool, and Danda Paudel.
\newblock A unified and interpretable emotion representation and expression generation.
\newblock In \emph{CVPR}, 2024.

\bibitem[Pexels(2024)]{pexels}
Pexels.
\newblock pexels.
\newblock \emph{https://www.pexels.com/}, 2024.

\bibitem[Ren et~al.(2021)Ren, Li, Chen, Li, and Liu]{ren2021pirenderer}
Yurui Ren, Ge~Li, Yuanqi Chen, Thomas~H Li, and Shan Liu.
\newblock Pirenderer: Controllable portrait image generation via semantic neural rendering.
\newblock In \emph{ICCV}, 2021.

\bibitem[Rombach et~al.(2022)Rombach, Blattmann, Lorenz, Esser, and Ommer]{rombach2022high}
Robin Rombach, Andreas Blattmann, Dominik Lorenz, Patrick Esser, and Bj{\"o}rn Ommer.
\newblock High-resolution image synthesis with latent diffusion models.
\newblock In \emph{CVPR}, pp.\  10684--10695, 2022.

\bibitem[Ruiz et~al.(2023)Ruiz, Li, Jampani, Pritch, Rubinstein, and Aberman]{ruiz2023dreambooth}
Nataniel Ruiz, Yuanzhen Li, Varun Jampani, Yael Pritch, Michael Rubinstein, and Kfir Aberman.
\newblock Dreambooth: Fine tuning text-to-image diffusion models for subject-driven generation.
\newblock In \emph{Proceedings of the IEEE/CVF conference on computer vision and pattern recognition}, pp.\  22500--22510, 2023.

\bibitem[Saharia et~al.(2022)Saharia, Chan, Saxena, Li, Whang, Denton, Ghasemipour, Gontijo~Lopes, Karagol~Ayan, Salimans, et~al.]{saharia2022photorealistic}
Chitwan Saharia, William Chan, Saurabh Saxena, Lala Li, Jay Whang, Emily~L Denton, Kamyar Ghasemipour, Raphael Gontijo~Lopes, Burcu Karagol~Ayan, Tim Salimans, et~al.
\newblock Photorealistic text-to-image diffusion models with deep language understanding.
\newblock \emph{NeurIPS}, 35:\penalty0 36479--36494, 2022.

\bibitem[Siarohin et~al.(2019{\natexlab{a}})Siarohin, Lathuili{\`e}re, Tulyakov, Ricci, and Sebe]{siarohin2019animating}
Aliaksandr Siarohin, St{\'e}phane Lathuili{\`e}re, Sergey Tulyakov, Elisa Ricci, and Nicu Sebe.
\newblock Animating arbitrary objects via deep motion transfer.
\newblock In \emph{CVPR}, 2019{\natexlab{a}}.

\bibitem[Siarohin et~al.(2019{\natexlab{b}})Siarohin, Lathuili{\`e}re, Tulyakov, Ricci, and Sebe]{siarohin2019first}
Aliaksandr Siarohin, St{\'e}phane Lathuili{\`e}re, Sergey Tulyakov, Elisa Ricci, and Nicu Sebe.
\newblock First order motion model for image animation.
\newblock \emph{NeurIPS}, 2019{\natexlab{b}}.

\bibitem[Simonyan \& Zisserman(2014)Simonyan and Zisserman]{simonyan2014very}
Karen Simonyan and Andrew Zisserman.
\newblock Very deep convolutional networks for large-scale image recognition.
\newblock \emph{arXiv preprint arXiv:1409.1556}, 2014.

\bibitem[Song et~al.(2020{\natexlab{a}})Song, Meng, and Ermon]{song2020denoising}
Jiaming Song, Chenlin Meng, and Stefano Ermon.
\newblock Denoising diffusion implicit models.
\newblock \emph{arXiv preprint arXiv:2010.02502}, 2020{\natexlab{a}}.

\bibitem[Song et~al.(2020{\natexlab{b}})Song, Sohl-Dickstein, Kingma, Kumar, Ermon, and Poole]{song2020score}
Yang Song, Jascha Sohl-Dickstein, Diederik~P Kingma, Abhishek Kumar, Stefano Ermon, and Ben Poole.
\newblock Score-based generative modeling through stochastic differential equations.
\newblock \emph{arXiv preprint arXiv:2011.13456}, 2020{\natexlab{b}}.

\bibitem[Tian et~al.(2024)Tian, Wang, Zhang, and Bo]{tian2024emo}
Linrui Tian, Qi~Wang, Bang Zhang, and Liefeng Bo.
\newblock Emo: Emote portrait alive-generating expressive portrait videos with audio2video diffusion model under weak conditions.
\newblock In \emph{ECCV}, 2024.

\bibitem[Tishby et~al.(2000)Tishby, Pereira, and Bialek]{tishby2000information}
Naftali Tishby, Fernando~C Pereira, and William Bialek.
\newblock The information bottleneck method.
\newblock \emph{arXiv preprint physics/0004057}, 2000.

\bibitem[Toisoul et~al.(2021)Toisoul, Kossaifi, Bulat, Tzimiropoulos, and Pantic]{toisoul2021estimation}
Antoine Toisoul, Jean Kossaifi, Adrian Bulat, Georgios Tzimiropoulos, and Maja Pantic.
\newblock Estimation of continuous valence and arousal levels from faces in naturalistic conditions.
\newblock \emph{Nature Machine Intelligence}, 2021.
\newblock URL \url{https://www.nature.com/articles/s42256-020-00280-0}.

\bibitem[Tseng et~al.(2022)Tseng, Castellon, and Liu]{tseng2022edge}
Jonathan Tseng, Rodrigo Castellon, and C.~Karen Liu.
\newblock Edge: Editable dance generation from music, 2022.

\bibitem[Varanka et~al.(2024)Varanka, Khor, Li, Wei, Kung, Sebe, and Zhao]{varanka2024towards}
Tuomas Varanka, Huai-Qian Khor, Yante Li, Mengting Wei, Hanwei Kung, Nicu Sebe, and Guoying Zhao.
\newblock Towards localized fine-grained control for facial expression generation.
\newblock \emph{arXiv preprint arXiv:2407.20175}, 2024.

\bibitem[Wang et~al.(2024)Wang, Tian, Zhang, Guan, Luo, Shen, Jiang, Gu, Han, and Yang]{wang2024v}
Cong Wang, Kuan Tian, Jun Zhang, Yonghang Guan, Feng Luo, Fei Shen, Zhiwei Jiang, Qing Gu, Xiao Han, and Wei Yang.
\newblock V-express: Conditional dropout for progressive training of portrait video generation.
\newblock \emph{arXiv preprint arXiv:2406.02511}, 2024.

\bibitem[Wang et~al.(2023)Wang, Deng, Yin, Shum, and Wang]{wang2023progressive}
Duomin Wang, Yu~Deng, Zixin Yin, Heung-Yeung Shum, and Baoyuan Wang.
\newblock Progressive disentangled representation learning for fine-grained controllable talking head synthesis.
\newblock In \emph{CVPR}, 2023.

\bibitem[Wang et~al.(2020)Wang, Wu, Song, Yang, Wu, Qian, He, Qiao, and Loy]{wang2020mead}
Kaisiyuan Wang, Qianyi Wu, Linsen Song, Zhuoqian Yang, Wayne Wu, Chen Qian, Ran He, Yu~Qiao, and Chen~Change Loy.
\newblock Mead: A large-scale audio-visual dataset for emotional talking-face generation.
\newblock In \emph{European Conference on Computer Vision}, pp.\  700--717. Springer, 2020.

\bibitem[Wang et~al.(2021)Wang, Mallya, and Liu]{wang2021one}
Ting-Chun Wang, Arun Mallya, and Ming-Yu Liu.
\newblock One-shot free-view neural talking-head synthesis for video conferencing.
\newblock In \emph{CVPR}, 2021.

\bibitem[Wang et~al.(2022)Wang, Yang, Bremond, and Dantcheva]{wang2022latent}
Yaohui Wang, Di~Yang, Francois Bremond, and Antitza Dantcheva.
\newblock Latent image animator: Learning to animate images via latent space navigation.
\newblock \emph{ICLR}, 2022.

\bibitem[Wei et~al.(2024)Wei, Yang, and Wang]{wei2024aniportrait}
Huawei Wei, Zejun Yang, and Zhisheng Wang.
\newblock Aniportrait: Audio-driven synthesis of photorealistic portrait animation.
\newblock \emph{arXiv preprint arXiv:2403.17694}, 2024.

\bibitem[Xie et~al.(2022)Xie, Wang, Zhang, Dong, and Shan]{xie2022vfhq}
Liangbin Xie, Xintao Wang, Honglun Zhang, Chao Dong, and Ying Shan.
\newblock Vfhq: A high-quality dataset and benchmark for video face super-resolution.
\newblock In \emph{The IEEE Conference on Computer Vision and Pattern Recognition Workshops (CVPRW)}, 2022.

\bibitem[Xie et~al.(2024)Xie, Xu, Song, Wang, Shi, and Luo]{xie2024x}
You Xie, Hongyi Xu, Guoxian Song, Chao Wang, Yichun Shi, and Linjie Luo.
\newblock X-portrait: Expressive portrait animation with hierarchical motion attention.
\newblock In \emph{SIGGRAPH}, 2024.

\bibitem[Xu et~al.(2024{\natexlab{a}})Xu, Liu, Xing, Wang, Sun, Dan, Huang, Li, Cheng, Tai, et~al.]{xu2024facechain}
Chao Xu, Yang Liu, Jiazheng Xing, Weida Wang, Mingze Sun, Jun Dan, Tianxin Huang, Siyuan Li, Zhi-Qi Cheng, Ying Tai, et~al.
\newblock Facechain-imagineid: Freely crafting high-fidelity diverse talking faces from disentangled audio.
\newblock In \emph{CVPR}, 2024{\natexlab{a}}.

\bibitem[Xu et~al.(2024{\natexlab{b}})Xu, Li, Su, Shang, Zhang, Liu, Wang, Van~Gool, Yao, and Zhu]{xu2024hallo}
Mingwang Xu, Hui Li, Qingkun Su, Hanlin Shang, Liwei Zhang, Ce~Liu, Jingdong Wang, Luc Van~Gool, Yao Yao, and Siyu Zhu.
\newblock Hallo: Hierarchical audio-driven visual synthesis for portrait image animation.
\newblock \emph{arXiv preprint arXiv:2406.08801}, 2024{\natexlab{b}}.

\bibitem[Xu et~al.(2024{\natexlab{c}})Xu, Chen, Guo, Yang, Li, Zang, Zhang, Tong, and Guo]{xu2024vasa}
Sicheng Xu, Guojun Chen, Yu-Xiao Guo, Jiaolong Yang, Chong Li, Zhenyu Zang, Yizhong Zhang, Xin Tong, and Baining Guo.
\newblock Vasa-1: Lifelike audio-driven talking faces generated in real time.
\newblock \emph{arXiv preprint arXiv:2404.10667}, 2024{\natexlab{c}}.

\bibitem[Xu et~al.(2024{\natexlab{d}})Xu, Zhang, Liew, Yan, Liu, Zhang, Feng, and Shou]{xu2024magicanimate}
Zhongcong Xu, Jianfeng Zhang, Jun~Hao Liew, Hanshu Yan, Jia-Wei Liu, Chenxu Zhang, Jiashi Feng, and Mike~Zheng Shou.
\newblock Magicanimate: Temporally consistent human image animation using diffusion model.
\newblock In \emph{CVPR}, 2024{\natexlab{d}}.

\bibitem[Yang et~al.(2024{\natexlab{a}})Yang, Li, Wu, Jing, Li, Ji, Liang, and Fan]{yang2024megactor}
Shurong Yang, Huadong Li, Juhao Wu, Minhao Jing, Linze Li, Renhe Ji, Jiajun Liang, and Haoqiang Fan.
\newblock Megactor: Harness the power of raw video for vivid portrait animation.
\newblock \emph{arXiv preprint arXiv:2405.20851}, 2024{\natexlab{a}}.

\bibitem[Yang et~al.(2024{\natexlab{b}})Yang, Teng, Zheng, Ding, Huang, Xu, Yang, Hong, Zhang, Feng, et~al.]{yang2024cogvideox}
Zhuoyi Yang, Jiayan Teng, Wendi Zheng, Ming Ding, Shiyu Huang, Jiazheng Xu, Yuanming Yang, Wenyi Hong, Xiaohan Zhang, Guanyu Feng, et~al.
\newblock Cogvideox: Text-to-video diffusion models with an expert transformer.
\newblock \emph{arXiv preprint arXiv:2408.06072}, 2024{\natexlab{b}}.

\bibitem[Yao et~al.(2021)Yao, Gholami, Shen, Mustafa, Keutzer, and Mahoney]{yao2021adahessian}
Zhewei Yao, Amir Gholami, Sheng Shen, Mustafa Mustafa, Kurt Keutzer, and Michael Mahoney.
\newblock Adahessian: An adaptive second order optimizer for machine learning.
\newblock In \emph{proceedings of the AAAI conference on artificial intelligence}, volume~35, pp.\  10665--10673, 2021.

\bibitem[Yin et~al.(2022)Yin, Zhang, Cun, Cao, Fan, Wang, Bai, Wu, Wang, and Yang]{yin2022styleheat}
Fei Yin, Yong Zhang, Xiaodong Cun, Mingdeng Cao, Yanbo Fan, Xuan Wang, Qingyan Bai, Baoyuan Wu, Jue Wang, and Yujiu Yang.
\newblock Styleheat: One-shot high-resolution editable talking face generation via pre-trained stylegan.
\newblock In \emph{ECCV}, 2022.

\bibitem[Zhang et~al.(2023{\natexlab{a}})Zhang, Zhang, Cun, Huang, Zhang, Zhao, Lu, and Shen]{zhang2023generating}
Jianrong Zhang, Yangsong Zhang, Xiaodong Cun, Shaoli Huang, Yong Zhang, Hongwei Zhao, Hongtao Lu, and Xi~Shen.
\newblock T2m-gpt: Generating human motion from textual descriptions with discrete representations.
\newblock In \emph{CVPR}, 2023{\natexlab{a}}.

\bibitem[Zhang et~al.(2023{\natexlab{b}})Zhang, Rao, and Agrawala]{zhang2023adding}
Lvmin Zhang, Anyi Rao, and Maneesh Agrawala.
\newblock Adding conditional control to text-to-image diffusion models.
\newblock In \emph{ICCV}, pp.\  3836--3847, 2023{\natexlab{b}}.

\bibitem[Zhang et~al.(2021)Zhang, Li, Ding, and Fan]{zhang2021flow}
Zhimeng Zhang, Lincheng Li, Yu~Ding, and Changjie Fan.
\newblock Flow-guided one-shot talking face generation with a high-resolution audio-visual dataset.
\newblock In \emph{CVPR}, pp.\  3661--3670, 2021.

\bibitem[Zhao \& Zhang(2022)Zhao and Zhang]{zhao2022thin}
Jian Zhao and Hui Zhang.
\newblock Thin-plate spline motion model for image animation.
\newblock In \emph{CVPR}, 2022.

\bibitem[Zheng et~al.(2024)Zheng, Peng, Yang, Shen, Li, Liu, Zhou, Li, and You]{opensora}
Zangwei Zheng, Xiangyu Peng, Tianji Yang, Chenhui Shen, Shenggui Li, Hongxin Liu, Yukun Zhou, Tianyi Li, and Yang You.
\newblock Open-sora: Democratizing efficient video production for all, March 2024.
\newblock URL \url{https://github.com/hpcaitech/Open-Sora}.

\bibitem[Zhou et~al.(2021)Zhou, Sun, Wu, Loy, Wang, and Liu]{zhou2021pose}
Hang Zhou, Yasheng Sun, Wayne Wu, Chen~Change Loy, Xiaogang Wang, and Ziwei Liu.
\newblock Pose-controllable talking face generation by implicitly modularized audio-visual representation.
\newblock In \emph{CVPR}, 2021.

\end{thebibliography}
